\documentclass[10pt,twocolumn,letterpaper]{article}

\usepackage{cvpr}
\usepackage{times}
\usepackage{epsfig}
\usepackage{graphicx}
\usepackage{amsmath}
\usepackage{amssymb}
 \usepackage{multirow}
 \usepackage{epstopdf}
 \usepackage{mathtools}
 \usepackage{tabularx}
 \usepackage{booktabs}
\usepackage{subcaption}
\usepackage{lipsum}
\usepackage{authblk}
\usepackage{url}

\newcommand{\tabincell}[2]{\begin{tabular}{@{}#1@{}}#2\end{tabular}} 
\newcolumntype{Y}{>{\centering\arraybackslash}X}

\usepackage[pagebackref=true,breaklinks=true,letterpaper=true,colorlinks,bookmarks=false]{hyperref}

\cvprfinalcopy 

\def\cvprPaperID{1291} 

\ifcvprfinal\fi
\begin{document}

\title{Pose2Seg: Detection Free Human Instance Segmentation}


\author{
Song-Hai Zhang$^{1,2}$, Ruilong Li$^{1,2}$, Xin Dong$^1$, Paul Rosin$^3$, Zixi Cai$^1$, Xi Han$^1$,  Dingcheng Yang$^1$, Haozhi Huang$^4$ and Shi-Min Hu$^{1,2}$ 
\vskip 0.3cm 
$^1$ Tsinghua University  \quad $^2$ BNRist \quad $^3$ Cardiff University   \quad  $^4$ Tencent AI Lab 
\vskip 0.2cm
{\tt\footnotesize \{shz, shimin\}@tsinghua.edu.cn, \{li-rl16, dong-x16, caizx15, x-han15, ydc15\}@mails.tsinghua.edu.cn, RosinPL@cardiff.ac.uk, matthzhuang@tencent.com
}
}

\maketitle

\begin{abstract}
The standard approach to image instance segmentation is to perform the object detection first, and then segment the object from the detection bounding-box. More recently, deep learning methods like Mask R-CNN~\cite{He2017Mask} perform them jointly. However, little research takes into account the uniqueness of the ``human'' category, which can be well defined by the pose skeleton. Moreover, the human pose skeleton can be used to better distinguish instances with heavy occlusion than using bounding-boxes. In this paper, we present a brand new pose-based instance segmentation framework\footnote{Codes are available: \href{https://github.com/liruilong940607/Pose2Seg}{https://github.com/liruilong940607/Pose2Seg}} for humans which separates instances based on human pose, rather than proposal region detection. We demonstrate that our pose-based framework can achieve better accuracy than the state-of-art detection-based approach on the human instance segmentation problem, and can moreover better handle occlusion.
Furthermore, there are few public datasets containing many heavily occluded humans along with comprehensive annotations, 
which makes this a challenging problem seldom noticed by researchers. 
Therefore, in this paper we introduce a new benchmark ``Occluded Human ({OCHuman})''\footnote{Dataset is available: \href{https://github.com/liruilong940607/OCHumanApi}{https://github.com/liruilong940607/OCHumanApi}}, which focuses on occluded humans with comprehensive annotations including bounding-box, human pose and instance masks.  This dataset contains 8110 detailed annotated human instances within 4731 images. With an average 0.67 {MaxIoU} for each person, {OCHuman} is the most complex and challenging dataset related to human instance segmentation. Through this dataset, we want to emphasize occlusion as a challenging problem for researchers to study.

\end{abstract}

\begin{figure}[t]
\centering
\includegraphics[width=\linewidth]{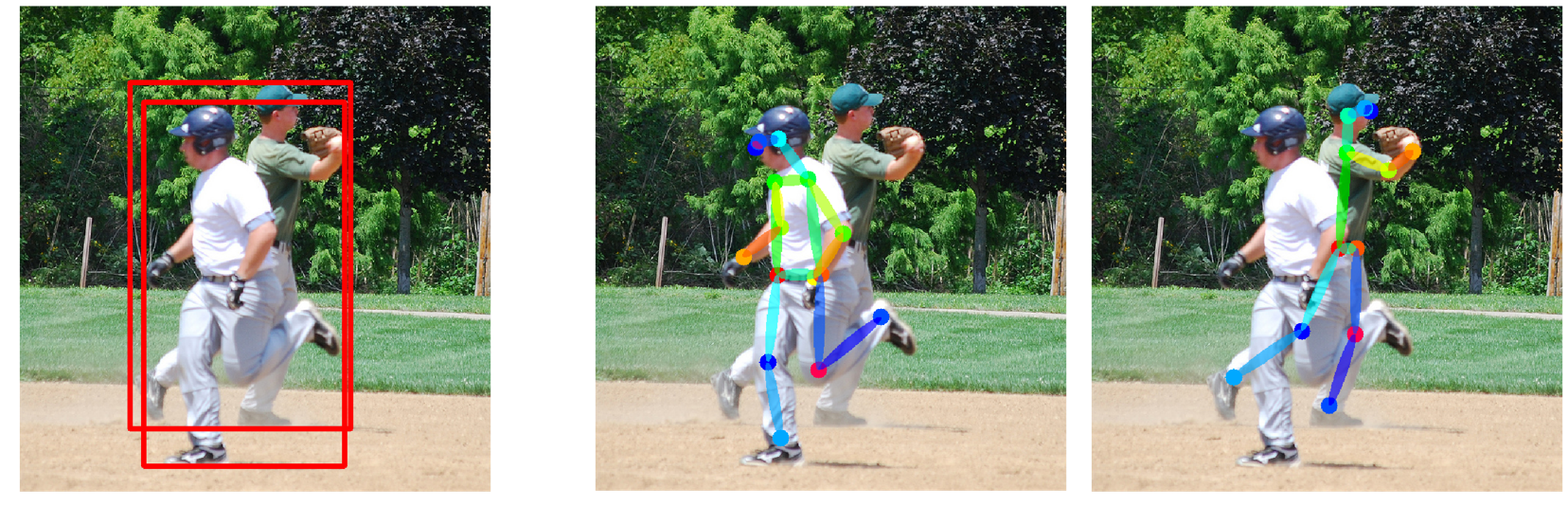}
\setlength{\abovecaptionskip}{-0.4cm}
\setlength{\belowcaptionskip}{-0.6cm}
\caption{Heavily occluded people are 
better separated using \emph{human pose} than using \emph{bounding-box}.}
\label{fig:improve_overlaps}
\end{figure}

\section{Introduction}
\label{intro}

In recent years, research related to ``humans'' in the computer vision community has become increasingly active because of the high demand for real-life applications. There has been much good research in the fields of human pose estimation~\cite{cao2017realtime,chen2017cascaded,fang2017rmpe,He2017Mask,lifkooee2019real,Newell2016Associative, xia2017survey}, pedestrian detection~\cite{mao2017can,zhang2016faster,zhang2018occluded}, portrait segmentation~\cite{shen2017high,shen2016automatic,Shen2016Deep}, and face recognition~\cite{ koestinger2011annotated, liu2017robust, ma2018robust, ouyang2016fast, wang2017joint, zhang2014facial, zhou2013extensive}, much of which has already produced practical value in real life.  This paper focuses on multi-person pose estimation and human instance segmentation, and proposes a pose-based human instance segmentation framework.

\emph{General Object Instance Segmentation} is a challenging problem which aims to predict pixel-level labels for each object instance in the image. Currently, those instance segmentation methods with highest accuracy~\cite{dai2016instance, He2017Mask, li2017fully,pinheiro2015learning} are all based on powerful \emph{object detection} baseline methods, such as Fast/Faster R-CNN~\cite{girshick2015fast,ren2015faster}, YOLO~\cite{redmon2016you}, which mostly follow a basic rule: first generate a large number of proposal regions, then remove the redundant regions using \emph{Non-maximum Suppression (NMS)}. However, when two objects of the same category have a large overlap, NMS will treat one of them as a redundant proposal region and eliminates it. This means that almost all the object detection methods cannot deal with the situation of large overlaps.
Moreover, even if the detection methods sometimes successfully detect two instances, the bounding-box is not suitable for instance segmentation in occluded cases. If two instances are heavily intertwined, they will both appear in the same bounding-box (like the case in Figure~\ref{fig:improve_overlaps}), which makes it hard for the segmentation network to identify which instance should be the target in this \emph{Region of Interest (RoI)}.

However, ``human'' is a special category in the computer vision community, and can be well defined by the pose skeleton. As shown in Figure~\ref{fig:improve_overlaps}, Human pose skeletons are more suitable for distinguishing two heavily intertwined people, because they can provide more distinct information about a person than bounding-boxes, such as the location and visibility of different body parts. \emph{ Multi-Person Pose Estimation} is also a very active topic in recent years, and there is already good progress \cite{cao2017realtime,chen2017cascaded,fang2017rmpe,huang2017coarse,Newell2016Associative,papandreou2017towards} on tackling this problem.
Although object detection methods are widely used by many multi-person pose estimation frameworks, some powerful bottom-up methods~\cite{cao2017realtime,Newell2016Associative} which do not rely on object detection also achieved good performance, including the \emph{COCO keypoints challenge 2016 winner}~\cite{cao2017realtime}. 
The main idea of the bottom-up methods is to first detect keypoints for each body part for all the people, and then group or connect those parts to form several instances of human pose, which makes it possible to seperate two intertwined human instances with a large overlap. Based on this observation, we present a new pose-based instance segmentation framework for humans which separates instances based on human pose rather than region proposal detection. Our pose-based framework works seamlessly with existing bottom-up pose estimation methods, and works better than the detection-based framework, especially in the case of occlusion.

\begin{figure}
\centering
\includegraphics[width=\linewidth]{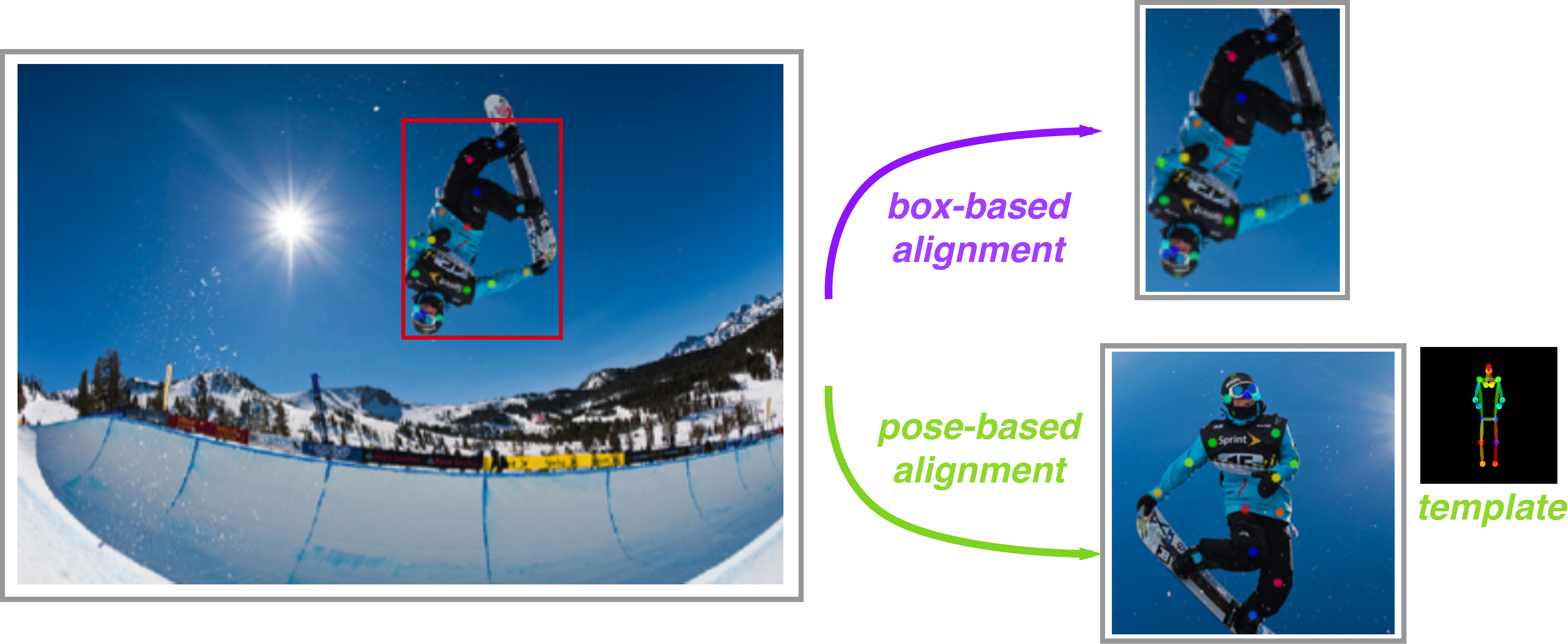}
\setlength{\abovecaptionskip}{-0.3cm}
\setlength{\belowcaptionskip}{-0.5cm}
\caption{Comparison of box-based alignment and our pose-based alignment (\emph{Affine-Align}). Objects with strange pose are corrected to a standard pose.}
\label{fig:skiing}
\end{figure}

Generally, there is an align module in the instance segmentation framework, for example, \emph{RoI-Align} in Mask R-CNN. The align module is used to crop the objects from the image using detection bounding boxes, and resize the objects to a uniform scale. Since it is hard to find a bounding box accurately from the object using human pose, we proposed an align module based on human pose, called \emph{Affine-Align}, which is a combination of scale, translation, rotation and left-right flip. An extra advantage of using \emph{Affine-Align} is that we can correct some objects with strange poses to a standard pose, like the inverted skiing human in Figure~\ref{fig:skiing}. 

Additionally, the human pose and human mask are not independent. Human pose can be approximately considered as a skeleton of the mask of the human instance. So we explicitly use human pose to guide the segmentation module by concatenating the \emph{Skeleton} features to the instance feature map after \emph{Affine-Align}. Our experiments demonstrate our \emph{Skeleton} features not only help to improve the accuracy of segmentation, but also give our network the ability to easily distinguish different instances that are heavily intertwined in the same RoI.

Severe occlusion between human bodies is often encountered in life, but current human-related public datasets either do not contain many severe occlusion situations~\cite{dollar2009pedestrian,geiger2012we,lin2014microsoft}, or lack comprehensive annotations of the human instances~\cite{shao2018crowdhuman}. Therefore, we introduce a new benchmark \emph{``Occluded Human (OCHuman)''} in this paper, which focuses on heavily occluded humans with comprehensive annotations including bounding-boxes, human poses and instance masks. This dataset contains 8110 detailed annotated human instances within 4731 images. On average, over 67\% of the bounding-box area of a human is occluded by one or several other persons, which makes this dataset the most complex and challenging dataset related to humans. Through this dataset, we want to emphasize occlusion as a challenging problem for researchers to study, and encourage current algorithms to become more practical for real life situations.

Our main contributions can be summarized as follows:

\begin{itemize}
\item We propose a brand new pose-based human instance segmentation framework which works better than the detection-based framework, especially in cases with occlusion.
\item We propose a pose-based align module, called \emph{Affine-Align}, which can align image windows into a uniform scale and direction based on human pose.
\item We explicitly use artific human \emph{Skeleton} features to guide the segmentation module and achieve a further improvement of the segmentation accuracy.
\item We introduce a new benchmark \emph{OCHuman} which focuses on the heavy occlusion problem, with comprehensive annotations including bounding-boxes, human poses and instance masks.
\end{itemize}

\begin{figure*}[t]
\centering
\includegraphics[width=\textwidth]{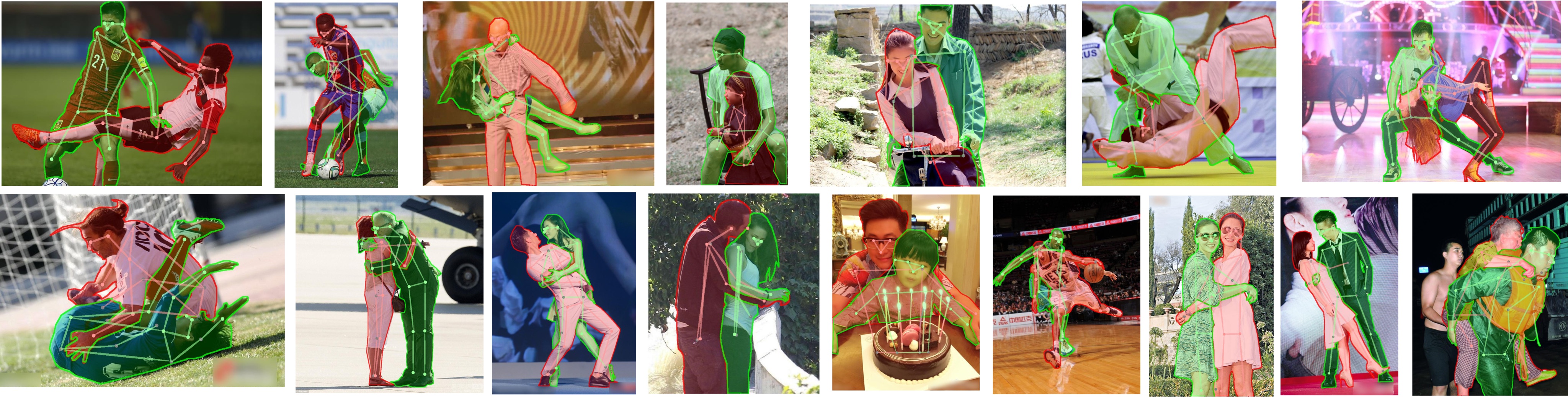} 
\setlength{\abovecaptionskip}{-0.3cm}
\setlength{\belowcaptionskip}{-0.5cm}
\caption{Samples of our \emph{OCHuman} dataset. All the annotated people in this dataset are heavily occluded with others, and have comprehensive annotations.}
\label{fig:dataset}
\end{figure*}

\section{Related Work}
\subsection{Multi-Person Pose Estimation}
Top-down methods~\cite{chen2017cascaded,fang2017rmpe,He2017Mask,huang2017coarse,papandreou2017towards} first employ object detection to crop each person, and then use a single-person pose estimation method on each human instance, which makes them all suffer from the defects of object detection methods on heavy occlusion.
While other bottom-up methods~\cite{cao2017realtime,insafutdinov2016deepercut,Newell2016Associative, rajchl2017deepcut} first detect body part keypoints of all the people, and then cluster these parts into instances of human pose.
Pishchulin \etal~\cite{rajchl2017deepcut} propose a complex framework of partitioning and labeling body-parts generated using a CNN. They solve the problem as an integer linear program, and jointly generate the detection and pose estimation results. Insafutdinov \etal~\cite{insafutdinov2016deepercut} use Resnet~\cite{he2016deep} to improve precision, and propose image-conditioned pairwise terms to increase speed. Cao \etal~\cite{cao2017realtime} use knowledge of the human structure, and predict a keypoints heatmap and PAFs, and finally connect the body parts. Newell \etal~\cite{Newell2016Associative} design a tag score map for each body part and use the score map to group body part keypoints.

\subsection{Instance Segmentation}
Some works~\cite{dai2015convolutional,girshick2015deformable,hariharan2014simultaneous,hariharan2015hypercolumns} employ a multi-stage pipeline which first uses detection to generate bounding boxes and then applies semantic segmentation. Others~\cite{dai2016instance,li2017fully,liu2017sgn,pinheiro2015learning} employ a tighter integration of detection and segmentation, e.g. jointly and  simultaneously performing detection and segmentation in an end-to-end framework~\cite{li2017fully}. Mask R-CNN~\cite{He2017Mask} is the state-of-art performing framework on the COCO~\cite{lin2014microsoft} dataset competition. 

\subsection{Harnessing Human Pose Estimation for Instance Segmentation}
There are three typical works that combine human pose estimation and instance segmentation. Mask R-CNN~\cite{He2017Mask} approach detects objects while generating instance segmentation and human pose estimation simultaneously in a single framework. But in their work, Mask R-CNN~\cite{He2017Mask} with mask-only performs better than combining keypoints and masks in the instance segmentation task. Pose2Instance~\cite{tripathi2017pose2instance} proposes a cascade network to harness human pose estimation for instance segmentation. Both of these two works rely on human detection, and perform poorly when two bounding boxes have a large overlap. 
More recently, PersonLab~\cite{papandreou2018personlab} treats instance segmentation as a pixel-wise clustering problem, and use human pose to refine the clustering results. Although their method is not based on bounding-box detection, they cannot perform as well as Mask R-CNN~\cite{He2017Mask} in the segmentation task.


\section{Occluded Human Benchmark}

Our \emph{``Occluded Human (OCHuman)''} dataset contains 8110 human instances within 4731 images. Each human instance is heavily occluded by one or several others. We use \emph{MaxIoU} to measure the severity of an object being occluded, which means the max \emph{IoU} with other same category objects in a single image. Those instances with MaxIoU $>$0.5 are referred to as \emph{heavy occlusion}, and are selected to form this dataset. Figure~\ref{fig:dataset} shows some samples from this dataset. With an average of 0.67 MaxIoU for each person, \emph{OCHuman} is the most challenging dataset related to human instances. Moreover, \emph{OCHuman} also has rich annotations. Each instance is annotated with a bounding-box for object detection, an instance binary mask for instance segmentation and 17 body joint locations for pose estimation. All images are collected from real-world scenarios containing people with challenging poses and viewpoints, various appearances and in a wide range of resolutions. With \emph{OCHuman}, we provide a new benchmark for the problem of \emph{occlusion}.

\subsection{Annotations}
For each image we first annotate the bounding-box of all humans present. Then we calculate the IoU between all the person pairs, and mark those persons with MaxIoU$>$0.5 as heavily occluded instances. Finally, we provide extra information for those occluded instance. The \emph{OCHuman} dataset contains three kinds of annotations related to humans: bounding-boxes, instance binary masks and 17 body joint locations. We reference the definition of body joints from~\cite{lin2014microsoft}, which are eye, nose, ear, shoulder, elbow, wrist, hip, knee and ankle. Except for the nose, all other joints have distinct left and right instances. 

\setlength{\tabcolsep}{10pt}
\begin{table}[t]
\small
\begin{center}
\begin{tabular}{lcc}
\toprule[1.5pt]
 & COCOPersons & OCHuman \\
 & (\emph{train+val}) & (\emph{val+test}) \\
\hline
\#images   & \bf{64115} & 4731  \\
\#persons  & \bf{273469}  & 8110  \\
\#persons ($oc_{0.5}$)    & 2619($<$1.0\%)  & $\bf{8110(100\%)}$  \\
\#persons ($oc_{0.75}$)  & 214($<$0.1\%)  & $\bf{2614(32\%)}$  \\
\#average MaxIoU 			   & $0.08$  & $\bf{0.67}$  \\
\bottomrule[1.5pt]
\end{tabular}
\end{center}
\setlength{\abovecaptionskip}{-0.1cm}
\setlength{\belowcaptionskip}{-0.5cm}
\caption{Comparison of different public datasets related to occluded human. ``persons ($oc_{X}$)'' represents occluded persons with MaxIoU $>$ X.}
\label{table:dataset_statistics}
\end{table}

\begin{figure*}[t]
\centering
\setlength{\abovecaptionskip}{0.1cm}
\setlength{\belowcaptionskip}{-0.5cm}
\includegraphics[width=\textwidth]{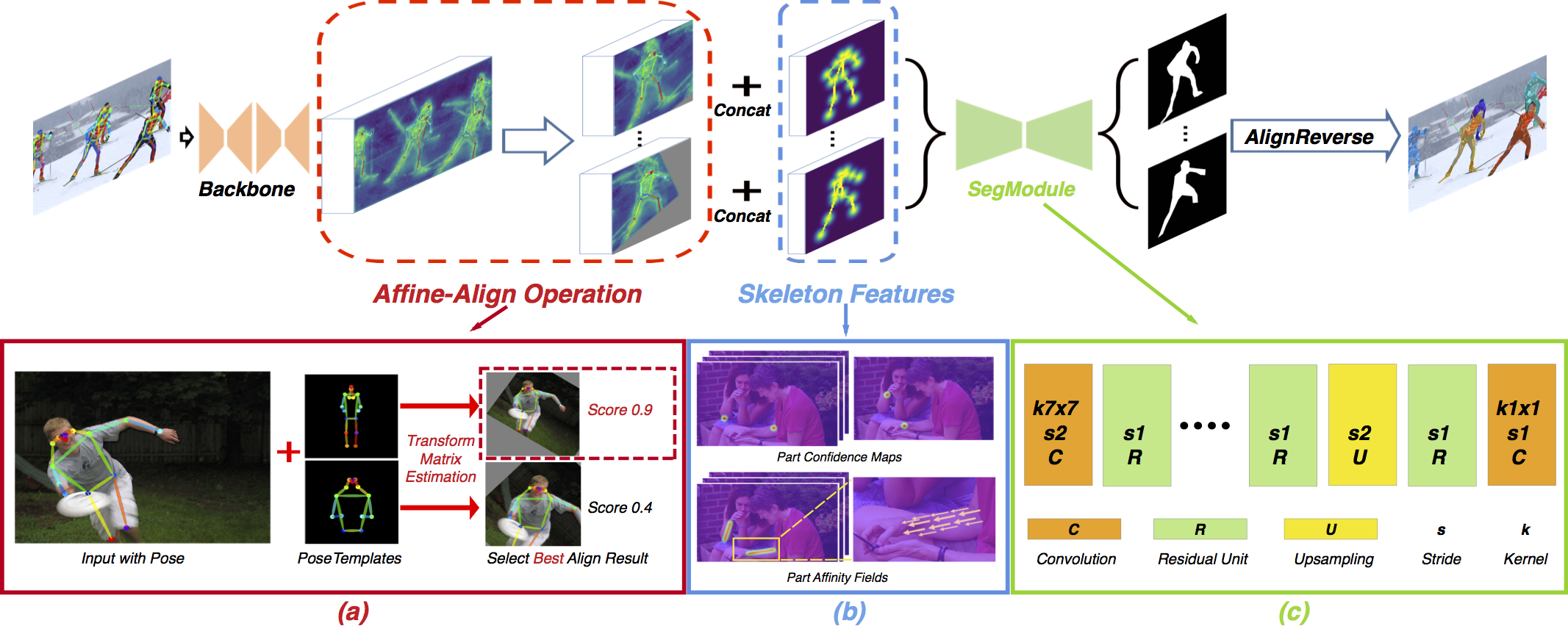} 
\caption{Overview of our network structure (Sec.~\ref{sec:overview}). (a) \emph{Affine-Align} operation (Sec.~\ref{sec:AffineAlign}). (b) Skeleton features (Sec.~\ref{sec:Skel}). (c) Structure of SegModule (Sec.~\ref{method:seg}), in which residual unit refers to~\cite{he2016deep}.}
\label{fig:structure}
\end{figure*}

\subsection{Dataset Splits}
\emph{OCHuman} dataset is designed for validation and testing. Since all the instances in this dataset are heavily occluded by other instances, we consider it is better to use general datasets such as COCO~\cite{lin2014microsoft} as a training set, then test the robustness of the segmentation methods to occlusion using this dataset, rather than performing training on only occluded cases. We split our dataset into separate validation and test sets. Following random selection, we arrive at a unique split consisting of 2500 validation and 2231 testing image, containing 4313 and 3797 instances respectively.
Furthermore, we divide instances in \emph{OCHuman} dataset into two subsets: \emph{OCHuman-Moderate} and \emph{OCHuman-Hard}. The first subset contains instances with MaxIoU in the range of 0.5 and 0.75, while the second contains instances with MaxIoU larger than 0.75, making it the more challenging subset. 
With these two subsets, we can evaluate the ability of algorithms to handle occlusions of different levels of severity.

\subsection{Dataset Statistics}

We compare our dataset with the person part of COCO in Table.~\ref{table:dataset_statistics}, which is currently the largest public dataset that contains both instance masks and human pose key-points. Although COCO includes comprehensive annotations, it contains few occluded human cases, and so this dataset cannot help to evaluate the capability of methods when faced with occlusion. \emph{OCHuman} is designed for all three most important tasks related to humans: detection, pose estimation and instance segmentation. It is the most challenging benchmark because of its heavy occlusion.


\section{Approach}

\subsection{Overview}
\label{sec:overview}
Our overall structure is shown in Figure~\ref{fig:structure}, which takes both the image and the human pose as input. Firstly, a base network is used to extract the features of the image. Then an align module, called \emph{Affine-Align}, is used to align \emph{RoIs} to a uniform size, which is $64\times64$ in this paper, based on the human pose. In the meantime, we generate \emph{Skeleton} features for each human instance and concatenate them to the \emph{RoIs}. Our segmentation module, which we called \emph{SegModule}, is designed based on the same residual unit in Resnet~\cite{he2016deep}. We carry out experiments on how the depth of \emph{SegModule} contributes to the performance of this system in Section~\ref{sec:SegModule}. Finally, we use the estimated matrices in \emph{Affine-Align} operation to reverse the alignment for each instance and get the final segmentation results. We describes our \emph{Affine-Align} operation, \emph{Skeleton} features and \emph{SegModule} in the following subsections.

\subsection{Affine-Align Operation}
\label{sec:AffineAlign}
Our \emph{Affine-Align} operation is inspired by the RoI-Pooling in Faster R-CNN~\cite{ren2015faster} and RoI-Align in Mask R-CNN~\cite{He2017Mask}. But unlike them, we align the people based on human pose instead of bounding-boxes. Specifically, as shown in Figure~\ref{fig:structure}(a), we first cluster the poses in the dataset and use the center of each cluster as pose templates, to represent the standard poses in the dataset. Then for each pose detected in the image, we estimate the affine transformation matrix $H$ between it and the templates, and chose the best $H$ based on the transformation error. Finally, we apply $H$ to the image or features and transform it to the desired resolution using bilinear interpolation. Details are introduced below.

\subsubsection{Human Pose Representation}
Human poses are represented as a list of vectors. Let vector $P = (C_1, C_2, ..., C_m) \in \mathbb{R}^{m \times 3} $  represent the pose of a single person, where $C_i = (x, y, v) \in \mathbb{R}^3 $ is a 3D vector 
representing the coordinates of a single part (such as right-shoulder, left-ankle) and the visibility of this body joint. \emph{m} is a dataset related parameter meaning the total number of parts in a single pose, which is 17 in COCO. 

\subsubsection{Pose Templates}
We cluster the pose templates from the training set to best represent the distribution of various human poses. We use K-means clustering~\cite{forgy1965cluster} to cluster the poses $(P_1,P_2,...,P_n)$ into $k(\leq n)$ sets $\mathbf{S}=\{S_1,S_2,...,S_k\}$ by optimizing Eq.~\ref{equ:kmeans1}, in which $P_{\mu i}$ is the mean of poses in $S_i$. We define the distance between two human poses using Eq.~\ref{equ:kmeans2} and Eq.~\ref{equ:kmeans3}, with several preprocessing steps: 
(1) We first crop a square-RoI of each instance using its bounding-box, and put the target into the center of the RoI, along with its pose coordinates.
(2) We resize this square-RoI to $1 \times 1$, so that the pose coordinates are all normalized to $(0, 1)$.
(3) We only count those poses which contain more than 8 valid points in the dataset to serve our purpose of creating the pose templates. Poses with few valid points cannot provide effective information and would act as outliers during K-means clustering.

\vspace{-0.3cm}
\begin{equation}
\small
\label{equ:kmeans1}
\mathop{\arg\min}_{S}\sum_{i=1}^K\sum_{P\in S_{i}} Dist(P, P_{\mu i})
\end{equation}

\begin{equation}
\small
\label{equ:kmeans2}
Dist(P, P_{\mu i})=\sum_{j=1}^m \|C_{j}-C_{\mu ij}\|^{2}
\end{equation}

\begin{equation}
\small
\label{equ:kmeans3}
C_{j} = 
\begin{cases*}
(x, y, 2) & if $C_{j}$ is visible  \\
(x, y, 1) & if $C_{j}$ is not visible\\
(0.5, 0.5, 0) & if $C_{j}$ is not in image 
\end{cases*} 
\end{equation}
\vspace{0.1cm}

\begin{figure}[t]
\centering
\includegraphics[width=\linewidth]{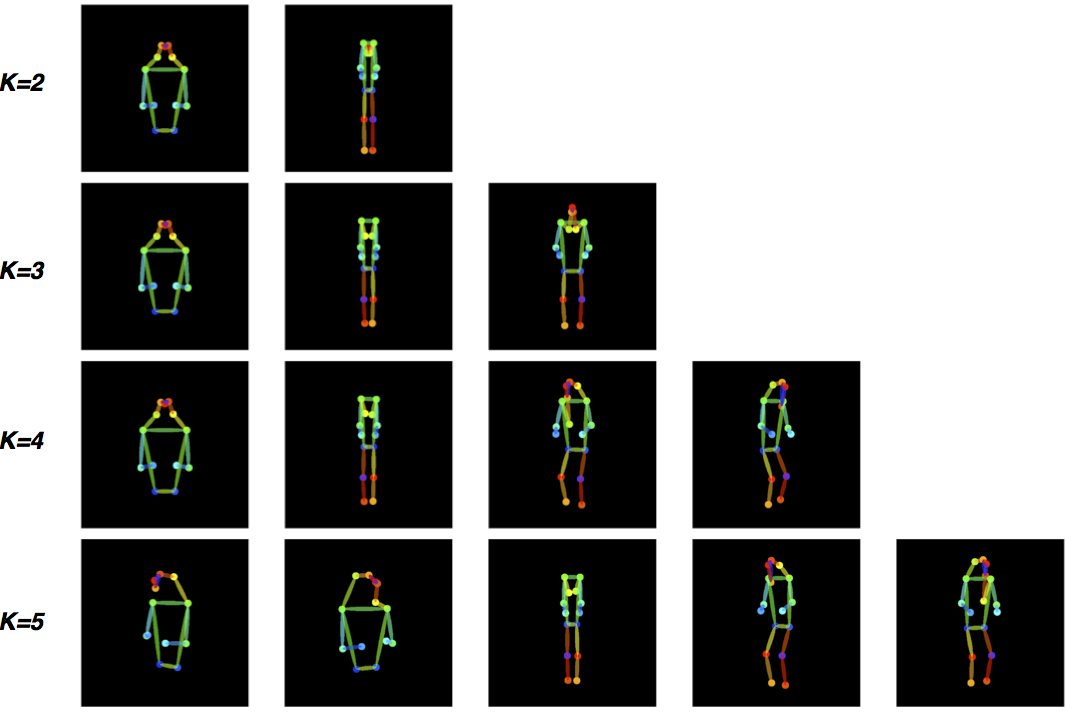} 
\setlength{\abovecaptionskip}{-0.3cm}
\setlength{\belowcaptionskip}{-0.5cm}
\caption{Pose templates clustered using K-means on COCO.}
\label{fig:templates}
\end{figure}

\noindent After K-means, we use the mean value of each set $P_{\mu i}$ to form the pose template and use it to represent the
whole group. We set those body joints with $v>0.5$ in $P_{\mu i}$ as valid points. Clustering results with different values of
K on the COCO training set are shown in Figure~\ref{fig:templates}. Although the results of K-means are heavily reliant on
initialization values, our multiple experimental results remain the same, which shows that there is a strong distinction between
different sets of human poses. After careful observation of those pose templates, we can find the two most frequent human poses
in COCO are a half-body pose and a full-body pose, which is in line with our common sense view of daily life. When $K=3$ in
K-means, we get a half-body pose, a full-body backview and a full-body frontview. When $K\geq 4$, the difference between left
and right are introduced. Since our align process copes with the left-right flip, $K\geq 4$ seems unnecessary for our framework. So finally, we choose $K=3$ to cluster pose templates in our approach.

\begin{figure*}[t]
\setlength{\abovecaptionskip}{0.1cm}
\setlength{\belowcaptionskip}{-0.4cm}
\centering
\includegraphics[width=\linewidth]{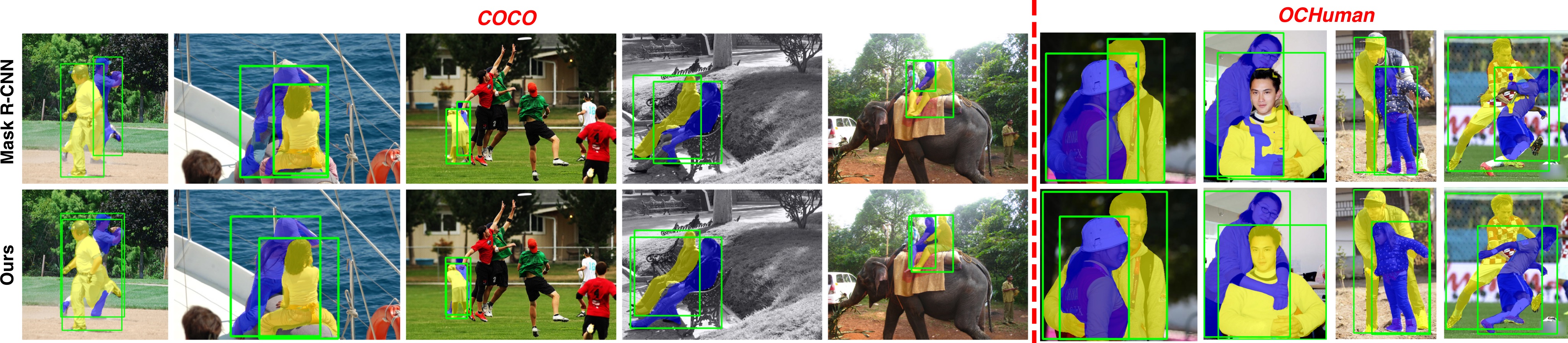}
\caption{Our method's results vs. Mask R-CNN~\cite{He2017Mask} on occlusion cases. Bounding-boxes in our results are generated using predicted masks for better visualization and comparison.}
\label{fig:combine_compare_overlaps}
\end{figure*}

\subsubsection{Estimate Affine Transformation Matrix}
Let vector $P_{\mu}$ represent a pose template, and $P$ represent a single person pose estimation result. We optimize
Eq.~\ref{equ:matrix_estimation1} to estimate an affine transformation matrix $H$ which transforms the pose coordinates to be as near as possible to the template coordinates. $H$ is a $2\times3$ matrix with 5 independent variables: rotation, scale factor, x-axis translation, y-axis translation and whether to do left-right flip. Since we have $K$ templates, we define a score for each $H^*$ based on the optimized error value, calculated by Eq.~\ref{equ:matrix_estimation2}, to choose the best template for each estimated pose, as shown in Figure~\ref{fig:structure}(a). In order to get the unique solution from Eq.~\ref{equ:matrix_estimation1}, $P_\mu$ and $P$ must contain at least three valid points in common, which can provide at least 6 independent equations for optimizing Eq.~\ref{equ:matrix_estimation1}. If none of our pose templates satisfy this condition, such as the case where there is only one valid point in $P$, the estimated transformation matrix $H^*$ will be calculated to align the whole image to the desired solution. In most case, this is reasonable because situations lacking valid points in the image mostly correspond to a single, large person in the image. 

\vspace{-0.4cm}
\begin{gather}
\small
\label{equ:matrix_estimation1}
H^{*} = \mathop{\arg\min}_{H} \ \| H\cdot P-P_{\mu} \|. \\
\label{equ:matrix_estimation2}
score = \exp( - \| H^{*}\cdot P - P_{\mu}) \| )
\end{gather}

\subsection{Skeleton Features}
\label{sec:Skel}
Figure~\ref{fig:structure}(b) shows our \emph{Skeleton} features. We adopt the part affinity fields (PAFs) from \cite{cao2017realtime}, which is a 2-channel vector field map for each skeleton. We use PAFs to represent the skeleton structure of a human pose. With 19 skeletons defined in the COCO dataset, PAFs is a 38-channel feature map for each human pose instance. We also use part confidence maps for body parts to emphasize the importance of those regions around the body part keypoints. For the COCO dataset, each human pose has a 17-channel part confidence map and a 38-channel PAFs map. So the total number of channels in our \emph{Skeleton} features is 55 for each human instance.

\subsection {SegModule}
\label{method:seg}
Since we introduced \emph{Skeleton} features after alignment to artificially extend the image features,  Our segmentation module, which we called \emph{SegModule}, needs to have enough receptive fields to not only fully understand these artificial features, but also learn the connections between them and the image features extracted by the base network. Therefore, we design \emph{SegModule} based on the resolution of the aligned RoIs. Figure~\ref{fig:structure}(c) demonstrates the overall architecture of our \emph{SegModule}. It starts with a $7\times7$, stride-2 convolution layer, and is followed by several standard residual units~\cite{he2016deep} to achieve
a large enough receptive field for the RoIs.
After that, a bilinear upsampling layer is used to restore the resolution, and another residual unit, along with a $1\times1$ convolution layer are used to predict the final result. 
Such a structure with 10 residual units can achieve about 50 pixels of receptive field, corresponding to our alignment size of $64\times64$. Fewer units will make the network less capable of learning, and more units enable little improvement on the learning ability. Table~\ref{table:segmodule} shows our experiment on this.


\section{Experiments}

\begin{figure*}[t]
\centering
\setlength{\abovecaptionskip}{0.1cm}
\setlength{\belowcaptionskip}{-0.3cm}
\includegraphics[width=\textwidth]{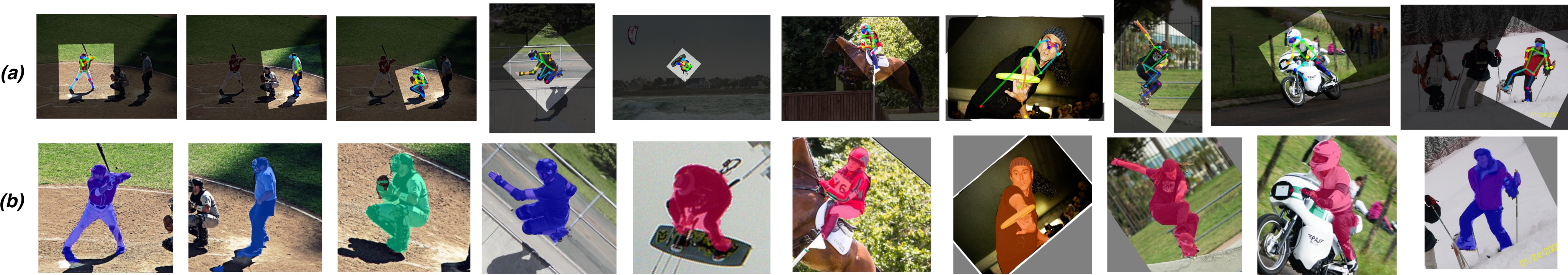}
\caption{More results of our \emph{Affine-Align} operation. (a) shows the align window on the original image. (b) shows the align results  and the segmentation results of our framework.}
\label{fig:affinealign_results}
\end{figure*}

\setlength{\tabcolsep}{6pt}
\begin{table*}[t]
\small
\begin{center}
\begin{subtable}[t]{0.45\linewidth}
\begin{tabular}{lllll}
\toprule[1.5pt]
\emph{Methods} & \emph{Backbone} & $AP$ & $AP_{M}$ & $AP_{H}$  \\
\hline\noalign{\smallskip}
Mask R-CNN & Resnet50-fpn & 0.163 & 0.194 & 0.113\\
\midrule[1pt]
\textbf{Ours} & Resnet50-fpn  & \textbf{0.222} & \textbf{0.261} & \textbf{0.150}\\
\textbf{Ours(GT Kpt)} & Resnet50-fpn  & \textbf{0.544} & \textbf{0.576} & \textbf{0.491}\\
\bottomrule[1.5pt]
\end{tabular}
\setlength{\abovecaptionskip}{0.0cm}
\caption{Performance on OCHuman \emph{val} set.}
\end{subtable}
\hspace{0.8cm}
\begin{subtable}[t]{0.45\linewidth}
\begin{tabular}{lllll}
\toprule[1.5pt]
\emph{Methods} & \emph{Backbone} & $AP$ & $AP_{M}$ & $AP_{H}$  \\
\hline\noalign{\smallskip}
Mask R-CNN & Resnet50-fpn & 0.169 & 0.189 & 0.128\\
\midrule[1pt]
\textbf{Ours} & Resnet50-fpn  & \textbf{0.238} & \textbf{0.266} & \textbf{0.175}\\
\textbf{Ours(GT Kpt)} & Resnet50-fpn  & \textbf{0.552} & \textbf{0.579} & \textbf{0.495}\\
\bottomrule[1.5pt]
\end{tabular}
\setlength{\abovecaptionskip}{0.0cm}
\caption{Performance on OCHuman \emph{test} set.}
\end{subtable}
\end{center}
\setlength{\abovecaptionskip}{-0.3cm}
\setlength{\belowcaptionskip}{-0.4cm}
\caption{Performance on occlusion. All methods are trained on COCOPersons train split, and tested on \emph{OCHuman}. Ours (GT Kpt) indicates our method with the input of ground-truth keypoints.}
\label{table:score_oc}
\end{table*}
\setlength{\tabcolsep}{1.4pt}

\setlength{\tabcolsep}{4pt}
\begin{table}
\small
\begin{center}
\begin{tabular}{llclclcl}
\toprule[1.5pt]
\emph{Methods} & \emph{Backbone} & $AP$ & $AP_M$ & $AP_L$ \\
\hline\noalign{\smallskip}
Mask R-CNN & Resnet50-fpn & 0.532 & 0.433 & 0.648\\
PersonLab & Resnet101 & - & 0.476 & 0.592\\
PersonLab & Resnet101(ms scale) & - & 0.492 & 0.621\\
PersonLab & Resnet152 & - & 0.483 & 0.595\\
PersonLab & Resnet152(ms scale) & -  & 0.497 & 0.621\\
\midrule[1pt]
\textbf{Ours} & Resnet50-fpn  & \textbf{0.555} & \bf{0.498} & \bf{0.670}\\
\textbf{Ours(GT Kpt)} & Resnet50-fpn  & \textbf{0.582}  & \bf{0.539} & \bf{0.679}\\
\bottomrule[1.5pt]
\end{tabular}
\end{center}
\setlength{\abovecaptionskip}{-0.1cm}
\setlength{\belowcaptionskip}{-0.5cm}
\caption{Performance on general cases. Mask R-CNN and ours are trained on the COCOPersons train split, and tested on the COCOPersons \textbf{val} split (without \emph{Small} category persons). Scores of PersonLab~\cite{papandreou2018personlab} is referred from their paper. Ours (GT Kpt) indicates our method with the input of ground-truth keypoints.}
\label{table:score_general}
\end{table}
\setlength{\tabcolsep}{1.4pt}

We evaluate our proposed method on two datasets: (1) \emph{OCHuman}, which is the largest validation dataset that is focused on heavily occluded humans, and proposed in this paper; and (2) \emph{COCOPersons} (the person category of COCO)~\cite{lin2014microsoft}, which contains the most common scenarios in daily life. Note that the \emph{Small} category persons in COCO is not contained in COCOPersons due to the lack of annotations of human pose. 

As far as we know, there are few public datasets which have labels for both human pose and human instance segmentation. COCO is the largest dataset that meets both of these requirements, so all of our models are trained end-to-end on the COCOPersons training set with the annotations of pose keypoints and segmentation masks. We compare our methods with Mask-RCNN~\cite{He2017Mask}, the well known detection based instance segmentation framework. For Mask-RCNN~\cite{He2017Mask}, we use the author's released code and configurations from \cite{Detectron2018}, and retrained and evaluate the model on the same dataset as our method. Our framework is implemented using Pytorch. The input resolution of our framework is 
$512 \times 512$ in all experiments. 
All our models are trained using the same training schedule, which is started by $learning rate=2e-4$, decayed by 0.1 after 33 epochs, and ended after 40 epochs. Each model is trained on a single TITAN X (Pascal) with $batchsize=4$ for 80 hours. No special techniques are used, such as iterative training, online hard-case mining, or multi-GPU synchronized batch normalization. Our method with images and keypoints as inputs can run about 20 FPS on a TITAX X (Pascal).

\subsection{Performance on occlusion}

In this experiment, we evaluate our method's capacity for handling occlusion cases compared with Mask-RCNN~\cite{He2017Mask} on the \emph{OCHuman} dataset. All methods in this experiment are trained on COCOPersons, including our keypoint detector baseline~\cite{Newell2016Associative} which achieves 0.285 / 0.303 AP on the keypoints task of \emph{OCHuman} \emph{val / test} set. As shown in Table~\ref{table:score_oc}, based on this keypoint detector baseline, our framework can achieve nearly 50\% higher than the performance of Mask R-CNN~\cite{He2017Mask} on this dataset. In addition, we test the upper limits of our pose-based framework using ground-truth (GT) keypoints as input, and more than double the accuracy. This demonstrates that with a better keypoint detector our framework can perform far better on occlusion problems.  Some results are shown in Figure~\ref{fig:combine_compare_overlaps}.

\subsection{Performance on general cases}
\begin{figure*}[t]
\setlength{\abovecaptionskip}{0.1cm}
\setlength{\belowcaptionskip}{-0.5cm}
\centering
\includegraphics[width=0.95\textwidth]{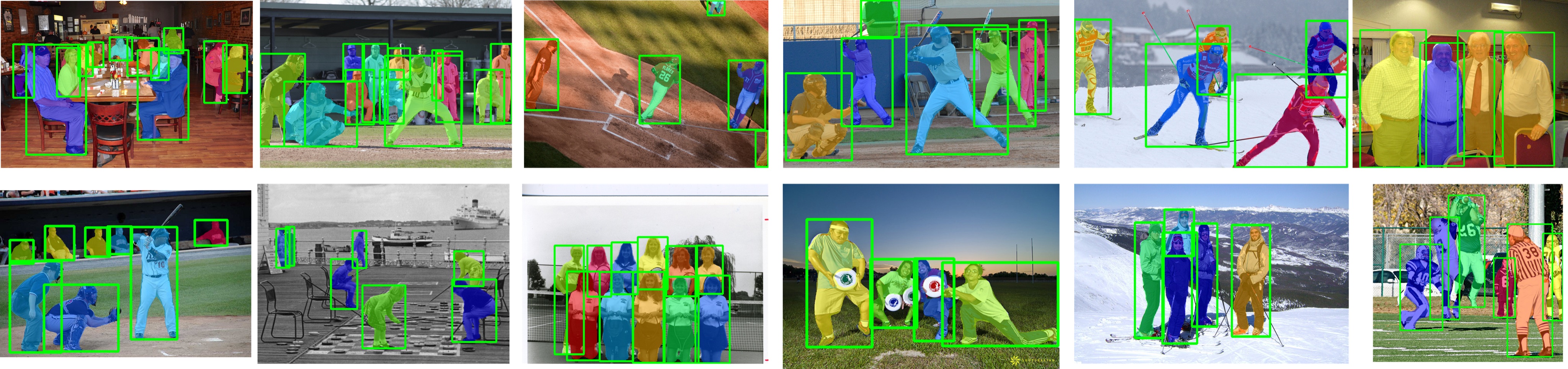}
\caption{More results of our instance segmentation framework on COCO. Bounding-boxes are generated using predicted masks for better visualization.}
\label{fig:final_results}
\end{figure*}

In this experiment, we evaluate our model on the \emph{COCOPerson} validation set using groundtruth keypoints as input, and get 0.582 AP on the instance segmentation task. We also evaluate the performance of our model under the predicted pose keypoints using our keypoint detector baseline~\cite{Newell2016Associative}, and achieve 0.555 AP. Mask R-CNN~\cite{He2017Mask} can only achieve 0.532 AP on this same dataset. We further compare our results with a recent work, PersonLab~\cite{papandreou2018personlab}. Scores of PersonLab~\cite{papandreou2018personlab} are taken from their paper, in which the detector is trained and tested on the whole person category of COCO. For fair comparison, we only compare against the results of the \emph{Median} and \emph{Large} categories. Our results surpass theirs with a heavier backbone and multi-scale prediction, as shown in Table~\ref{table:score_general}. Figure~\ref{fig:final_results} and Figure~\ref{fig:affinealign_results} show some results of our instance segmentation framework and our \emph{Affine-Align} operation, respectively.

\subsection{Ablation Experiments}

\subsubsection{Affine-Align v.s. RoI-Align}

\paragraph{Occluded Cases} In this experiment, we replace the align module in our framework with RoI-Align based on groundtruth (Gt) bounding-box, and re-train our model with nothing else changed. As shown in Table~\ref{table:noSkel}, this box-based alignment strategy can achieve 0.476 AP on \emph{OCHuman} validation set. Our \emph{Affine-Align} based on Gt human pose can achieve 0.544 AP on this same dataset. This means that even if we do not take into account the NMS's deficiencies on handling occlusion (which is eliminated by using GT bounding-boxes), the box-based alignment strategy still does not perform as well as our pose-based alignment strategy in the instance segmentation task of occlusion. The reason is that rotation is allowed in \emph{Affine-Align}, which helps to better distinguish two heavy intertwined people by aligning into discriminative RoIs. Strong discrimination RoIs are essential for the segmentation network to  locate and extract the specific target. 

\setlength{\tabcolsep}{8pt}
\begin{table}[t]
\small
\begin{center}
\begin{tabular}{ccccc}
\toprule[1.5pt]
\emph{N} Residual Units & 5 & 10 & 15 & 20 \\
\midrule[0.5pt]
Receptive Field   & $\sim30$ & $\sim50$ & $\sim70$ & $\sim90$  \\
\midrule[0.5pt]
AP   & 0.545 & 0.555 & 0.555 & 0.556  \\
\bottomrule[1.5pt]
\end{tabular}
\end{center}
\setlength{\abovecaptionskip}{-0.1cm}
\setlength{\belowcaptionskip}{-0.5cm}
\caption{Experiments on the depth of SegModule under $64\times64$ RoIs. 10 residual units with a receptive field of about 50 pixels is enough for this alignment size. Deeper architecture brings little benefits. All scores are tested on the COCOPerson \emph{val} set.}
\label{table:segmodule}
\end{table}

\vspace{-0.3cm}
\paragraph{General Cases} We also experiment on COCOPerson validation set. If we allow using both Gt bounding-box and Gt keypoint as input, the best performance is achieved by combining RoI-Align and our \emph{Skeleton} features (0.648 AP).  While simultaneously requiring bounding-box and keypoint as input is a rather strict requirement, and both of them can introduce error to the framework when using predicted results instead of ground-truth. If we contraint the framework to only rely on one of them, combining \emph{Affine-Align} with \emph{Skeleton} features can achieved better performance than using RoI-Align stategy on COCOPerson (0.582 AP v.s. 0.568 AP). What's more, the upper limits of the box-based framework is limited by NMS, especially in the case of occlusion. In comparison, our pose-based alignment strategy has no such limits.
\vspace{-0.3cm}
\paragraph{Intuitive Pose-based Alignment}  An intuitive idea of pose-based alignment is to first generate bounding-boxes based on human pose key-points, and then use a box-based alignment strategy, such as RoI-Align, to align each person into a RoI. We take the maximum and minimum values of the valid key-points as the generated bounding-box, and expand the generated bounding-box by a factor $\alpha$ to simulate the accurate bounding-box as much as possible. We treat $\alpha$ as a hyperparameter and search for the best value during testing. Table~\ref{table:noSkel} shows that no matter how this hyperparameter $\alpha$ is adjusted, the performance still cannot match our \emph{Affine-Align} strategy.

\subsubsection{With/Without Skeleton Features}

We also experiment on the contribution of our artificial \emph{Skeleton} features. Table~\ref{table:noSkel} shows that our \emph{Skeleton} features are good for different kinds of align strategies because manually concatenating the features of human pose can explicitly provide more information for the network, and lead to a more accurate result. This is more effective for situations where there is more than one person in the RoIs (which is very common), because \emph{Skeleton} features can explicitly guide the network to focus on the specific person. Also, due to this component our framework can better segment the person under occlusion than the previous methods.

\subsubsection{SegModule}
\label{sec:SegModule}
We have discussed in Section~\ref{method:seg} that the receptive field is an important factor to be considered in designing the SegModule. So we experiment how the receptive field of SegModule affects our system. 
We achieve different receptive fields by stacking different numbers of residual units after the first convolution.
Besides that, all the other components stay unchanged. As shown in Table~\ref{table:segmodule}, our SegModule with 10 residual units can achieve about 50 pixels of receptive field, which is enough for our $64\times64$ alignment size. A large enough receptive field can provide enough learning ability to understand the image features and artificial features globally. Fewer units will make the network less capable of learning, and more units have little help with the learning ability,

\setlength{\tabcolsep}{1.4pt}
\begin{table}[t]
\footnotesize
\begin{center}
\setlength{\belowcaptionskip}{0pt}
\renewcommand\arraystretch{1.2}
\begin{tabularx}{\linewidth}{|Y|Y|>{\centering\arraybackslash}p{0.9cm}|Y|Y|}
\hline
\tabincell{c}{Training\\Method} & \tabincell{c}{Testing\\Method} & \tabincell{c}{BBox\\Expand} & \tabincell{c}{AP\\\emph{(OCHuman)}} & \tabincell{c}{AP\\\emph{(COCOPerson)}} \\
\hline
\multirow{9}*{\tabincell{c}{GT BBOX\\ + RoI-Align \\ (+/-) Skeleton}} & \tabincell{c}{GT BBOX\\ + RoI-Align \\ (+/-) Skeleton} & {\bf ---}& {\bf \tabincell{c}{0.476*/0.133}} & {\bf \tabincell{c}{0.648*/0.568}}   \\
\cline{2-5}
& \multirow{8}*{\tabincell{c}{GT KPT \\to BBOX\\ + RoI-Align \\ (+/-) Skeleton}} & 30\% &  \tabincell{c}{0.436/{\bf 0.124}}&  \tabincell{c}{0.431/0.354}   \\
\cline{3-5}
& &  40\% &  \tabincell{c}{{\bf 0.441}/0.115}&  \tabincell{c}{0.460/0.372}    \\
\cline{3-5}
& &  50\% &  \tabincell{c}{0.437/0.104}&  \tabincell{c}{0.477/{\bf 0.380}}    \\
\cline{3-5}
& & 60\% &  \tabincell{c}{0.429/0.093}&  \tabincell{c}{0.489/0.379}    \\
\cline{3-5}
& & 70\% &  \tabincell{c}{0.420/0.083}&  \tabincell{c}{0.497/0.371}    \\
\cline{3-5}
& & 80\% &  \tabincell{c}{0.411/0.074} &  \tabincell{c}{{\bf 0.501}/0.357}    \\
\cline{3-5}
& & 90\% &   \tabincell{c}{0.403/0.065} &  \tabincell{c}{0.500/0.343}    \\
\cline{3-5}
& & 100\% &   \tabincell{c}{0.393/0.057} &  \tabincell{c}{0.500/0.325}    \\
\hline
\tabincell{c}{GT KPT\\ + Affine-Align} & \tabincell{c}{GT KPT\\ + Affine-Align}  & {\bf ---} & {\bf 0.544/0.141} & {\bf 0.582/0.386} \\
\hline
\end{tabularx}
\end{center}
\setlength{\abovecaptionskip}{-0.1cm}
\setlength{\belowcaptionskip}{-0.4cm}
\caption{Ablation experiments on OCHuman \emph{val} set and COCOPerson \emph{val} set about different \emph{alignment} strategies and \emph{Skeleton} features. All scores are tested using ground-truth (GT) bounding-box (BBOX) or keypoint (KPT). `GT KPT to BBOX' represents taking the maximum and minimum values of the valid KPT as the BBOX, and expanding the BBOX by a factor. Notice that scores marked by * rely on both BBOX and KPT as input, while others rely on only one of them.}
\label{table:noSkel}
\end{table}

\section{Conclusion}

In this paper, we propose a pose-based human instance segmentation framework. We designed an Affine-Align operation for
selecting RoIs based on pose instead of bounding-boxes. We explicitly concatenate the human pose skeleton feature to the image
feature in the network to further improve the performance. Compared with the traditional detection based instance segmentation
framework, our pose-based system can achieve a better performance in the general case, and can moreover better handling
occlusion. In addition, we introduce a new dataset called \emph{OCHuman}, which focuses on heavily occluded humans, as a challenging benchmark on occlusion problem.

\vspace{-0.3cm}
{ \paragraph{ Acknowledgement:} This work was supported by the Natural Science Foundation of China (61772298, 61521002), Research Grant of Beijing Higher Institution Engineering Research Center and Tsinghua-Tencent Joint Laboratory for Internet Innovation Technology.}

{\small
\bibliographystyle{ieee}
\bibliography{egbib}

\begin{thebibliography}{10}\itemsep=-1pt

\bibitem{cao2017realtime}
Zhe Cao, Tomas Simon, Shih-En Wei, and Yaser Sheikh.
\newblock Realtime multi-person 2d pose estimation using part affinity fields.
\newblock In {\em Proceedings of the IEEE Conference on Computer Vision and
  Pattern Recognition}, pages 7291--7299, 2017.

\bibitem{chen2017cascaded}
Yilun Chen, Zhicheng Wang, Yuxiang Peng, Zhiqiang Zhang, Gang Yu, and Jian Sun.
\newblock Cascaded pyramid network for multi-person pose estimation.
\newblock {\em arXiv preprint arXiv:1711.07319}, 2017.

\bibitem{dai2016instance}
Jifeng Dai, Kaiming He, Yi Li, Shaoqing Ren, and Jian Sun.
\newblock Instance-sensitive fully convolutional networks.
\newblock In {\em European Conference on Computer Vision}, pages 534--549.
  Springer, 2016.

\bibitem{dai2015convolutional}
Jifeng Dai, Kaiming He, and Jian Sun.
\newblock Convolutional feature masking for joint object and stuff
  segmentation.
\newblock In {\em Proceedings of the IEEE Conference on Computer Vision and
  Pattern Recognition}, pages 3992--4000, 2015.

\bibitem{dollar2009pedestrian}
Piotr Doll{\'a}r, Christian Wojek, Bernt Schiele, and Pietro Perona.
\newblock Pedestrian detection: A benchmark.
\newblock In {\em Computer Vision and Pattern Recognition, 2009. CVPR 2009.
  IEEE Conference on}, pages 304--311. IEEE, 2009.

\bibitem{fang2017rmpe}
Hao-Shu Fang, Shuqin Xie, Yu-Wing Tai, and Cewu Lu.
\newblock {RMPE}: Regional multi-person pose estimation.
\newblock In {\em Proceedings of the IEEE Conference on Computer Vision and
  Pattern Recognition}, pages 2334--2343, 2017.

\bibitem{forgy1965cluster}
Edward~W Forgy.
\newblock Cluster analysis of multivariate data: efficiency versus
  interpretability of classifications.
\newblock {\em biometrics}, 21:768--769, 1965.

\bibitem{geiger2012we}
Andreas Geiger, Philip Lenz, and Raquel Urtasun.
\newblock Are we ready for autonomous driving? the kitti vision benchmark
  suite.
\newblock In {\em Computer Vision and Pattern Recognition (CVPR), 2012 IEEE
  Conference on}, pages 3354--3361. IEEE, 2012.

\bibitem{girshick2015fast}
Ross Girshick.
\newblock Fast r-cnn.
\newblock In {\em Proceedings of the IEEE international conference on computer
  vision}, pages 1440--1448, 2015.

\bibitem{girshick2015deformable}
Ross Girshick, Forrest Iandola, Trevor Darrell, and Jitendra Malik.
\newblock Deformable part models are convolutional neural networks.
\newblock In {\em Proceedings of the IEEE conference on Computer Vision and
  Pattern Recognition}, pages 437--446, 2015.

\bibitem{Detectron2018}
Ross Girshick, Ilija Radosavovic, Georgia Gkioxari, Piotr Doll\'{a}r, and
  Kaiming He.
\newblock Detectron.
\newblock \url{https://github.com/facebookresearch/detectron}, 2018.

\bibitem{hariharan2014simultaneous}
Bharath Hariharan, Pablo Arbel{\'a}ez, Ross Girshick, and Jitendra Malik.
\newblock Simultaneous detection and segmentation.
\newblock In {\em European Conference on Computer Vision}, pages 297--312.
  Springer, 2014.

\bibitem{hariharan2015hypercolumns}
Bharath Hariharan, Pablo Arbel{\'a}ez, Ross Girshick, and Jitendra Malik.
\newblock Hypercolumns for object segmentation and fine-grained localization.
\newblock In {\em Proceedings of the IEEE conference on computer vision and
  pattern recognition}, pages 447--456, 2015.

\bibitem{He2017Mask}
Kaiming He, Georgia Gkioxari, Piotr Doll{\'a}r, and Ross Girshick.
\newblock Mask r-cnn.
\newblock In {\em Computer Vision (ICCV), 2017 IEEE International Conference
  on}, pages 2980--2988. IEEE, 2017.

\bibitem{he2016deep}
Kaiming He, Xiangyu Zhang, Shaoqing Ren, and Jian Sun.
\newblock Deep residual learning for image recognition.
\newblock In {\em Proceedings of the IEEE conference on computer vision and
  pattern recognition}, pages 770--778, 2016.

\bibitem{huang2017coarse}
Shaoli Huang, Mingming Gong, and Dacheng Tao.
\newblock A coarse-fine network for keypoint localization.
\newblock In {\em The IEEE International Conference on Computer Vision (ICCV)},
  volume~2, 2017.

\bibitem{insafutdinov2016deepercut}
Eldar Insafutdinov, Leonid Pishchulin, Bjoern Andres, Mykhaylo Andriluka, and
  Bernt Schiele.
\newblock {DeeperCut}: A deeper, stronger, and faster multi-person pose
  estimation model.
\newblock In {\em European Conference on Computer Vision}, pages 34--50.
  Springer, 2016.

\bibitem{koestinger2011annotated}
Martin Koestinger, Paul Wohlhart, Peter~M Roth, and Horst Bischof.
\newblock Annotated facial landmarks in the wild: A large-scale, real-world
  database for facial landmark localization.
\newblock In {\em 2011 IEEE international conference on computer vision
  workshops (ICCV workshops)}, pages 2144--2151. IEEE, 2011.

\bibitem{li2017fully}
Yi Li, Haozhi Qi, Jifeng Dai, Xiangyang Ji, and Yichen Wei.
\newblock Fully convolutional instance-aware semantic segmentation.
\newblock In {\em IEEE Conf. on Computer Vision and Pattern Recognition
  (CVPR)}, pages 2359--2367, 2017.

\bibitem{lifkooee2019real}
Masoud~Zadghorban Lifkooee, Celong Liu, Yongqing Liang, Yimin Zhu, and Xin Li.
\newblock Real-time avatar pose transfer and motion generation using locally
  encoded laplacian offsets.
\newblock {\em Journal of Computer Science and Technology}, 34(2):256--271,
  2019.

\bibitem{lin2014microsoft}
Tsung-Yi Lin, Michael Maire, Serge Belongie, James Hays, Pietro Perona, Deva
  Ramanan, Piotr Doll{\'a}r, and C~Lawrence Zitnick.
\newblock Microsoft coco: Common objects in context.
\newblock In {\em European conference on computer vision}, pages 740--755.
  Springer, 2014.

\bibitem{liu2017sgn}
Shu Liu, Jiaya Jia, Sanja Fidler, and Raquel Urtasun.
\newblock Sgn: Sequential grouping networks for instance segmentation.
\newblock In {\em The IEEE International Conference on Computer Vision (ICCV)},
  2017.

\bibitem{liu2017robust}
Shuang Liu, Yongqiang Zhang, Xiaosong Yang, Daming Shi, and Jian~J Zhang.
\newblock Robust facial landmark detection and tracking across poses and
  expressions for in-the-wild monocular video.
\newblock {\em Computational Visual Media}, 3(1):33--47, 2017.

\bibitem{ma2018robust}
Xiao Ma, Fandong Zhang, Yuelong Li, and Jufu Feng.
\newblock Robust sparse representation based face recognition in an adaptive
  weighted spatial pyramid structure.
\newblock {\em Science China Information Sciences}, 61(1):012101, 2018.

\bibitem{mao2017can}
Jiayuan Mao, Tete Xiao, Yuning Jiang, and Zhimin Cao.
\newblock What can help pedestrian detection?
\newblock In {\em Proceedings of the IEEE Conference on Computer Vision and
  Pattern Recognition}, pages 3127--3136, 2017.

\bibitem{Newell2016Associative}
Alejandro Newell, Zhiao Huang, and Jia Deng.
\newblock Associative embedding: End-to-end learning for joint detection and
  grouping.
\newblock In {\em Advances in Neural Information Processing Systems}, pages
  2277--2287, 2017.

\bibitem{ouyang2016fast}
Peng Ouyang, Shouyi Yin, Chenchen Deng, Leibo Liu, and Shaojun Wei.
\newblock A fast face detection architecture for auto-focus in smart-phones and
  digital cameras.
\newblock {\em Science China Information Sciences}, 59(12):122402, 2016.

\bibitem{papandreou2018personlab}
George Papandreou, Tyler Zhu, Liang-Chieh Chen, Spyros Gidaris, Jonathan
  Tompson, and Kevin Murphy.
\newblock Personlab: Person pose estimation and instance segmentation with a
  bottom-up, part-based, geometric embedding model.
\newblock {\em arXiv preprint arXiv:1803.08225}, 2018.

\bibitem{papandreou2017towards}
George Papandreou, Tyler Zhu, Nori Kanazawa, Alexander Toshev, Jonathan
  Tompson, Chris Bregler, and Kevin Murphy.
\newblock Towards accurate multi-person pose estimation in the wild.
\newblock In {\em Proceedings of the IEEE Conference on Computer Vision and
  Pattern Recognition}, pages 4903--4911, 2017.

\bibitem{pinheiro2015learning}
Pedro~O Pinheiro, Ronan Collobert, and Piotr Doll{\'a}r.
\newblock Learning to segment object candidates.
\newblock In {\em Advances in Neural Information Processing Systems}, pages
  1990--1998, 2015.

\bibitem{rajchl2017deepcut}
Martin Rajchl, Matthew~CH Lee, Ozan Oktay, Konstantinos Kamnitsas, Jonathan
  Passerat-Palmbach, Wenjia Bai, Mellisa Damodaram, Mary~A Rutherford, Joseph~V
  Hajnal, Bernhard Kainz, et~al.
\newblock Deepcut: Object segmentation from bounding box annotations using
  convolutional neural networks.
\newblock {\em IEEE Transactions on Medical Imaging}, 36(2):674--683, 2017.

\bibitem{redmon2016you}
Joseph Redmon, Santosh Divvala, Ross Girshick, and Ali Farhadi.
\newblock You only look once: Unified, real-time object detection.
\newblock In {\em Proceedings of the IEEE conference on computer vision and
  pattern recognition}, pages 779--788, 2016.

\bibitem{ren2015faster}
Shaoqing Ren, Kaiming He, Ross Girshick, and Jian Sun.
\newblock Faster r-cnn: Towards real-time object detection with region proposal
  networks.
\newblock In {\em Advances in neural information processing systems}, pages
  91--99, 2015.

\bibitem{shao2018crowdhuman}
Shuai Shao, Zijian Zhao, Boxun Li, Tete Xiao, Gang Yu, Xiangyu Zhang, and Jian
  Sun.
\newblock Crowdhuman: A benchmark for detecting human in a crowd.
\newblock {\em arXiv preprint arXiv:1805.00123}, 2018.

\bibitem{shen2017high}
Xiaoyong Shen, Hongyun Gao, Xin Tao, Chao Zhou, and Jiaya Jia.
\newblock High-quality correspondence and segmentation estimation for dual-lens
  smart-phone portraits.
\newblock In {\em Proceedings of the IEEE International Conference on Computer
  Vision}, pages 3257--3266, 2017.

\bibitem{shen2016automatic}
Xiaoyong Shen, Aaron Hertzmann, Jiaya Jia, Sylvain Paris, Brian Price, Eli
  Shechtman, and Ian Sachs.
\newblock Automatic portrait segmentation for image stylization.
\newblock In {\em Computer Graphics Forum}, volume~35, pages 93--102. Wiley
  Online Library, 2016.

\bibitem{Shen2016Deep}
Xiaoyong Shen, Tao Xin, Hongyun Gao, Zhou Chao, and Jiaya Jia.
\newblock Deep automatic portrait matting.
\newblock In {\em European Conference on Computer Vision}, 2016.

\bibitem{tripathi2017pose2instance}
Subarna Tripathi, Maxwell Collins, Matthew Brown, and Serge Belongie.
\newblock Pose2instance: Harnessing keypoints for person instance segmentation.
\newblock {\em arXiv preprint arXiv:1704.01152}, 2017.

\bibitem{wang2017joint}
Jie Wang, Juyong Zhang, Changwei Luo, and Falai Chen.
\newblock Joint head pose and facial landmark regression from depth images.
\newblock {\em Computational Visual Media}, 3(3):229--241, 2017.

\bibitem{xia2017survey}
Shihong Xia, Lin Gao, Yu-Kun Lai, Ming-Ze Yuan, and Jinxiang Chai.
\newblock A survey on human performance capture and animation.
\newblock {\em Journal of Computer Science and Technology}, 32(3):536--554,
  2017.

\bibitem{zhang2016faster}
Liliang Zhang, Liang Lin, Xiaodan Liang, and Kaiming He.
\newblock Is faster r-cnn doing well for pedestrian detection?
\newblock In {\em European conference on computer vision}, pages 443--457.
  Springer, 2016.

\bibitem{zhang2018occluded}
Shanshan Zhang, Jian Yang, and Bernt Schiele.
\newblock Occluded pedestrian detection through guided attention in cnns.
\newblock In {\em Proceedings of the IEEE Conference on Computer Vision and
  Pattern Recognition}, pages 6995--7003, 2018.

\bibitem{zhang2014facial}
Zhanpeng Zhang, Ping Luo, Chen~Change Loy, and Xiaoou Tang.
\newblock Facial landmark detection by deep multi-task learning.
\newblock In {\em European conference on computer vision}, pages 94--108.
  Springer, 2014.

\bibitem{zhou2013extensive}
Erjin Zhou, Haoqiang Fan, Zhimin Cao, Yuning Jiang, and Qi Yin.
\newblock Extensive facial landmark localization with coarse-to-fine
  convolutional network cascade.
\newblock In {\em Proceedings of the IEEE International Conference on Computer
  Vision Workshops}, pages 386--391, 2013.

\end{thebibliography}
}

\end{document}